\documentclass[conference]{IEEEtran}
\usepackage{stmaryrd}
\usepackage{amsfonts}
\usepackage{multicol,lipsum}
\usepackage{amsmath,epsfig,amssymb,subfigure,bm,dsfont}
\usepackage{latexsym}
\usepackage{dcolumn}
\usepackage{amsmath}
\usepackage{epsf}
\usepackage{float}
\usepackage{textcomp}
\usepackage[bookmarks=false]{hyperref}
\usepackage{CJK}
\usepackage{times}
\usepackage{mathptmx}
\newcommand\blfootnote[1]{%
\begingroup
\renewcommand\thefootnote{}\footnote{#1}%
\addtocounter{footnote}{-1}%
\endgroup
}
\makeatletter

\usepackage{graphicx,times,epsfig,amsmath} 
\usepackage{balance,amssymb}
\usepackage[mathscr]{eucal}

\graphicspath{{figs/},{figs/faces/}}

\IEEEoverridecommandlockouts    

\begin{document}

\title{\ \\ \LARGE\bf  An Analysis on the Learning Rules of the Skip-Gram Model \thanks{Canlin Zhang is with the Department of Mathematics, Florida State University, Tallahassee, Florida 32306, USA (email: czhang@math.fsu.edu).} \thanks{Xiuwen Liu and Daniel Bi\'s are with the Department of Computer Science, Florida State University, Tallahassee, Florida 32306, USA (email: liux@cs.fsu.edu;  bis@cs.fsu.edu).}}

\author{Canlin Zhang, Xiuwen Liu and Daniel Bi\'s}


\maketitle

\begin{abstract}
  To improve the generalization of the representations for natural language processing tasks, words are commonly represented using vectors, where distances among the vectors are related to the similarity of the words. While word2vec, the state-of-the-art implementation of the skip-gram model, is widely used and improves the performance of many natural language processing tasks, its mechanism is not yet well understood. 
  
In this work, we derive the learning rules for the skip-gram model and establish their close relationship to competitive learning. In addition, we provide the global optimal solution constraints for the skip-gram model and validate them by experimental results.
  
\end{abstract}

\blfootnote{\textcopyright 2019 IEEE. This paper is published in 2019 International Joint Conference on Neural Networks (IJCNN). DOI: 10.1109/IJCNN.2019.8852182. Personal use of this material is permitted. Permission from IEEE must be obtained for all other uses, in any current or future media, including reprinting/republishing this material for advertising or promotional purposes, creating new collective works, for resale or redistribution to servers or lists, or reuse of any copyrighted component of this work in other works.}

\section{Introduction}

\PARstart{I}{n} the last few years, performance on natural language
processing tasks has improved significantly due to the application of deep learning architectures and better representations of words \cite{bengio2003} \cite{huang2012} \cite{oja89}.
To improve generalization and reduce the complexity of language models, words are commonly represented
using dense vectors, where similar vectors represent similar words \cite{mikolov2} \cite{mikolov3}. Among many such representations,
word2vec (short for Word-to-Vector) is widely used due to its computational efficiency and its ability to capture interesting
analogue relationships \cite{mikolov1} \cite{mikolov4}. In addition, systems built on word2vec representations often lead to significant
performance improvements. 

However, it is not well understood why word2vec exhibits these desirable properties. Although many researchers intended to find the source of efficiency of the word2vec, many works did not provide strict mathematical analysis on the formulas of the skip-gram model. The main contribution of this work is three-fold: First, we examine the gradient formulas of the skip-gram model and then derive the underlying learning rules for both the input and output vectors. Second, we establish that word2vec leads to a competitive learning rule \cite{shinozaki2018} for word representations. Third, given the training corpus, we provide the global optimal solution constraints on both the input and output vectors in the skip-gram model.

The paper is organized as follows: In Section II, we present the skip-gram model as well as word2vec for learning vector representations of words.
The learning rules of the skip-gram model as well as its connections to competitive learning are shown in section III. In Section IV, the global optimal solution constraints on the vectors of the skip-gram model are proved first; then, experimental results obtained on both a toy dataset and a big training corpus are provided to support our results. After that, the connections between the learning rules and the global optimal solution constraints are discussed. Finally, Section V concludes the paper with a brief summary and discussions on our future work.\\

\section{The Skip-Gram Model for Vector Representations of Words}
The skip-gram model implemented by word2vec \cite{mikolov1} can be described like this:

Suppose we are given a length-$T$ text training corpus $\{w_1,\cdots, w_T\}$. Based on this corpus, we can build a dictionary including $W$ words: $\mathcal{D}=\{w_1,\cdots, w_W\}$, where the words in dictionary $\mathcal{D}$ are descended according to their frequencies in the training corpus.

 Then, for each word $w$ in dictionary $\mathcal{D}$, two vector representations are provided: the $\mathbf{input \ vector}$ ($\mathbf{or \ word}$ $\mathbf{ embedding}$) $v_w$ and the $\mathbf{output \ vector}$ ($\mathbf{or \ context}$ $\mathbf{embedding}$) $v_w'$, both of which are initialized via random normal distribution \cite{initialize}. Based on the embedding vectors, the conditional probability $p(w_O|w_I)$ between any two words $w_I$ and $w_O$ in $\mathcal{D}$ are estimated using the Softmax function:
\begin{align}
\hat{p}(w_O|w_I)=\frac{\exp({v_{w_O}'}^Tv_{w_I})}{\sum_{w=1}^W\exp({v_{w}'}^Tv_{w_I})},
\end{align}
where ${v'}^T$ means the transformation of vector $v'$, and ${v'}^Tv$ means the inner product between the two vectors $v'$ and $v$.

Therefore, the goal of the skip-gram model is to maximize:
\begin{align}
E = \frac{1}{T}\sum_{t=1}^T\sum_{-c\leq j \leq c, j\neq 0}\log \hat{p}(w_{t+j}|w_t).
\end{align}
where $c$ is the radius of the center-removed context window at $w_t$.

Note that we use $\hat{p}$ rather than $p$ to represent the probability estimation given by the vectors, which differs from the typical notification in the papers on skip-gram model. In the following sections, we will do analysis on both the estimated probability $\hat{p}$ given by the vectors and the ground-truth probability $p$ based on the training corpus. Hence, we differ these two concepts ahead.  

Since the calculation of formula (1) requires $\mathcal{O}(W)$ inner products and exponentials, researchers usually do not directly use it in practice. Instead, many simplified versions of formula (1) are applied \cite{mnih2009} \cite{mnih2005}. But, Mikolov et al. \cite{mikolov1} came up with an efficient and effective method to approximate $\hat{p}(w_O|w_I)$: Instead of calculating $\sum_{w=1}^V\exp({v_{w}'}^Tv_{w_I})$ throught the words in the vocabulary, $\sum_{k=1}^K\exp({v_{w_k}'}^Tv_{w_I})$ is used as an approximation with randomly chosen words $\{w_1,w_2,\cdots,w_K\}$ from the distribution $P(w)$, where $K$ is around 2-5 for big datasets and 5-20 for small ones. The best specific distribution $P(w)$ is the unigram distribution $U(w)$ raised to the 3/4rd power, i.e., $P(w)=U(w)^{3/4}/Z$.

 Then, in order to maximize $\log \hat{p}(w_O|w_I)$, one only needs to maximize ${v_{w_O}'}^Tv_{w_I}$ and minimize $\sum_{k=1}^K\exp({v_{w_k}'}^Tv_{w_I})$, which is why the words $w_1,w_2,\cdots,w_K$ are called negative samples. Moreover, the exponential function is usually replaced by the sigmoid function $\sigma(x)=\frac{1}{1+\exp(-x)}$ in practice to avoid underflow. That is, Mikolov et al. aim at maximizing
\begin{align}
\log\sigma({v_{w_O}'}^Tv_{w_I})+\sum_{k=1}^K\mathbb{E}_{w_k\sim P(w)}\log\sigma(-{v_{w_k}'}^Tv_{w_I}),
\end{align}
for $w_I=w_t$ and $w_O=w_{t+j}$ in formula (2).

Formula (3) applying on word embeddings is the first efficient language model for neural network training. This skip-gram model implemented with negative sampling (SGNS) provides surprisingly meaningful results. Models using the SGNS pre-trained word embeddings provide not only good performance on many NLP tasks, but also a series of interesting analogue relationships \cite{mikolov5}. 

On the other hand, according to formula (3), it is not hard to see that the method of negative sampling itself ``involves no magic": it only provides a simple and efficient way to approximate the conditional probability $\hat{p}(w_{t+j}|w_t)$. We claim that the efficiency of the SGNS model lies in the skip-gram algorithm and the usage of dense vector embeddings of words instead of one-hot vectors. By implementing formula (3), the skip-gram model makes the embedding vectors of words with similar contexts converge to each other in the vector space. Hence, semantic information and analogue relationships are captured by the vector distribution. Detailed explanations are provided in the following two sections.

\section{The Learning Rules for the Skip-Gram Model}
In this section, we shall provide a systematic understanding on the learning rules of the skip-gram model. We shall first find the gradient formula for each input and output vector, based on which the connections between the skip-gram model and the competitive learning will be addressed. 

We reform the average log probability $E$ as:
\begin{align*}
  &  E  =    \frac{1}{T}\sum_{t=1}^{T}\sum_{\mbox{\scriptsize$\begin{array}{c}-c\le j \le c, j \ne 0
\end{array}$}}\log \hat{p}(w_{t+j}|w_t) \\
  & =  \frac{1}{T}\sum_{t=1}^{T}\!\!\sum_{\mbox{\scriptsize$\begin{array}{c}-c\le j \le c, j \ne 0
\end{array}$}}\!\!
  \left(v_{w_{t+j}}'^Tv_{w_t}-\log\left(\sum_{w=1}^{W}\exp(v_w'^T v_{w_t})\right)\!\!\right)\\
  & =  \frac{1}{T}\!\sum_{t=1}^{T}\!
  \left(\!\!\!\left(\!\!\!\!\!\!\sum_{\mbox{\scriptsize$\begin{array}{c}-c\le j \le c, j \ne 0
\end{array}$}}\!\!\!\!\!v_{w_{t+j}}'^Tv_{w_t}\!\!\!\right)\!-2c\log\!\left(\sum_{w=1}^{W}\exp(v_w'^T v_{w_t})\!\!\right)\!\!\!\right)
\end{align*}

Given a fixed word $w_s$ in the dictionary, the gradient of its input vector $v_{w_s}$ will be:
\begin{align*}
 & \frac{\partial E}{\partial v_{w_{s,i}}} =
  \frac{1}{T}\!\!\!\!\sum_{\mbox{\scriptsize$\begin{array}{c}t=1,\\
w_t=w_s 
\end{array}$}}^T\!\!\!\!\!\!\!\sum_{\mbox{\scriptsize$\begin{array}{c}-c\le j \le c,\\
 j \ne 0
\end{array}$}}\!\!\!\!
  \left(\!v_{w_{t+j,i}}'-\frac{\sum_{w=1}^{W}\exp(v_{w}'^T v_{w_t}) v_{w_i}'}
       {\sum_{w=1}^{W}\exp(v_{w}'^T v_{w_t})}\!\right) \\
       & =
       \frac{1}{T}\!\!\!\!\sum_{\mbox{\scriptsize$\begin{array}{c}t=1,\\
w_t=w_s 
\end{array}$}}^T\!\!\!\!\!\!\!\sum_{\mbox{\scriptsize$\begin{array}{c}-c\le j \le c,\\
 j \ne 0
\end{array}$}}\!\!\!\!
  \left(\!v_{w_{t+j,i}}'-\sum_{w=1}^{W}\frac{\exp(v_{w}'^T v_{w_s})}
       {\sum_{\tilde{w}=1}^{W}\exp(v_{\tilde{w}}'^T v_{w_s})}v_{w_i}'\right)\\
       & = \frac{1}{T}\!\!\sum_{\mbox{\scriptsize$\begin{array}{c}t=1,\\
w_t=w_s 
\end{array}$}}^T\!\!\!\!\!\!\!\sum_{\mbox{\scriptsize$\begin{array}{c}-c\le j \le c,\\
 j \ne 0
\end{array}$}}\!\!\!\!
        \left(v_{w_{t+j,i}}'-\sum_{w=1}^{W}\hat{p}(w|w_s) v_{w_i}'\right).
\end{align*}

Hence, the gradient formula for the entire input vector $v_{w_s}$ will be:
\begin{align*}
& \frac{\partial E}{\partial v_{w_{s}}} = \frac{1}{T}\!\!\sum_{\mbox{\scriptsize$\begin{array}{c}t=1,\\
w_t=w_s 
\end{array}$}}^T\!\!\!\!\!\!\!\sum_{\mbox{\scriptsize$\begin{array}{c}-c\le j \le c,\\
 j \ne 0
\end{array}$}}\!\!\!\!
        \left(v_{w_{t+j}}'-\sum_{w=1}^{W}\hat{p}(w|w_s) v_{w}'\right)\\
        & = \frac{1}{T}\!\!\sum_{\mbox{\scriptsize$\begin{array}{c}t=1,\\
w_t=w_s 
\end{array}$}}^T\!\!\!\!\!\!\!\sum_{\mbox{\scriptsize$\begin{array}{c}-c\le j \le c,\\
 j \ne 0, w_{t+j}=w
\end{array}$}}\!\!\!\!
        \left(v_{w}'-\sum_{\tilde{w}=1}^{W}\hat{p}(\tilde{w}|w_s) v_{\tilde{w}}'\right)\\
        & = \frac{1}{T}\!\!\sum_{\mbox{\scriptsize$\begin{array}{c}t=1,\\
w_t=w_s 
\end{array}$}}^T\!\!\!\!\!\!\!\sum_{\mbox{\scriptsize$\begin{array}{c}-c\le j \le c,\\
 j \ne 0,w_{t+j}=w
\end{array}$}}\!\!\!\!\!\!\!\!\!
        \left(\!\!\!\!\Big(\!\!1\!-\!\hat{p}(w|w_s)\!\!\Big)v_{w}'-\!\!\!\!\!\!\!\!\sum_{\mbox{\scriptsize$\begin{array}{c}\tilde{w}=1, \tilde{w}\neq w
\end{array}$}}^{W}\!\!\!\!\!\!\!\!\hat{p}(\tilde{w}|w_s) v_{\tilde{w}}'\!\!\right)
\end{align*}
where $w_{t+j}=w$ means that the word appearing at the position $t+j$ in the training corpus is the word $w$ in the dictionary.

Note that the training purpose of the skip-gram model is to maximize the average log probability $E$. Yet in practice, researchers always apply gradient descent to minimize $-E$ due to programming facts. However, in order to provide a clear theoretical analysis, we directly apply $\mathbf{gradient \ ascent}$ with respect to $\partial E/\partial v_{w_s}$ to maximize $E$ in our learning rule. Hence, the learning rule for updating the input vector $v_{w_s}$ will be:
\begin{align}
v_{w_{s}}^{\text{new}}\!\! = v_{w_{s}}^{\text{old}} + \frac{\eta}{T}\!\!\!\!\!\!\!\!\sum_{\mbox{\scriptsize$\begin{array}{c}t=1,\\
w_t=w_s 
\end{array}$}}^T\!\!\!\!\!\!\!\!\!\!\sum_{\mbox{\scriptsize$\begin{array}{c}-c\le j \le c,\\
 j \ne 0,\\
w_{t+j}=w
\end{array}$}}\!\!\!\!\!\!\!
        \left(\!\!\!\!\Big(\!\!1\!-\!\hat{p}(w|w_s)\!\Big)\!v_{w}'-\!\!\!\!\!\!\!\!\!\sum_{\mbox{\scriptsize$\begin{array}{c}\tilde{w}=1,\\
 \tilde{w}\neq w
\end{array}$}}^{W}\!\!\!\!\!\!\!\hat{p}(\tilde{w}|w_s) v_{\tilde{w}}'\!\!\!\right)
\end{align}
where $\eta$ is the learning rate.

 Intuitively speaking, adding a vector $\vec{b}$ to the vector $\vec{a}$ will make $\vec{a}$ move towards the direction of $\vec{b}$, or make the angle between $\vec{a}$ and $\vec{b}$ smaller. On the contrast, subtracting vector $\vec{b}$ from the vector $\vec{a}$ (i.e. $\vec{a}-\vec{b}$) will make $\vec{a}$ move away from $\vec{b}$, or make the angle between the two vectors larger. 
 
By analyzing terms in the large bracket of formula (4), one can see that the term $(1-\hat{p}(w|w_s))$ as well as each $\hat{p}(\tilde{w}|w_s)$ is always positive since $0<\hat{p}(w|w_s)<1$ for any word $w$. Hence, intuitively speaking, the vector $(1-\hat{p}(w|w_s))v_{w}'$ is added to $v_{w_s}$, while vectors $\hat{p}(\tilde{w}|w_s)v_{\tilde{w}}'$ for all the $\tilde{w}\neq w$ are subtracted from $v_{w_s}$. This means that the gradient ascent will make the input vector $v_{w_s}$ move towards the output vector $v_{w}'$ of word $w$ that appears in the context window of word $w_s$. Meanwhile, the gradient ascent will make $v_{w_s}$ move away from all the output vectors $v_{\tilde{w}}'$ other than $v_{w}'$. 

This process is a form of competitive learning: If a word $w$ appears in the context of the word $w_s$, it shall compete against all the other words to ``pull" the input vector $v_{w_s}$ closer to its own output vector $v_{w}'$. But we need to indicate that there is a major difference between the gradient ascent of the skip-gram model and the typical winner-takes-all (WTA) algorithm of competitive learning: In the back propagation of WTA, the gradients to all the loser neurons are zero, which means that ``the winner takes all while the losers stand still" \cite{shinozaki2018}. In the gradient ascent of the skip-gram model, however, the losers (which are all the words $\tilde{w}$ other than $w$) will be even worse: They have to ``push" the input vector $v_{w_s}$ away from their own output vectors $v_{\tilde{w}}'$. That is, the competitive learning rule for updating an input vector in the skip-gram model is ``the winner takes its winning while the losers even lose more".

The implementation of SGNS is also compatible with our analysis here: Maximizing formula (3) leads to maximizing the inner product $v_{w_O}'^Tv_{w_I}$ while minimizing the inner products $v_{w_k}'^Tv_{w_I}$ for all the words $w_k$. This means that the input vector $v_{w_I}$ of the word $w_I$ will be pulled closer towards the output vector $v_{w_O}'$, if word $w_O$ is in the context of word $w_I$; And meanwhile, $v_{w_I}$ will be pushed away from the output vectors $v_{w_k}'$, where $w_k$ are randomly chosen words (negative samples). That is, the negative samples play the role of simulating all the ``loser" words $\tilde{w}\neq w_O$: Since computing $\hat{p}(\tilde{w}|w_I)$ for all the words $\tilde{w}$ other than $w_O$ is not feasible, SGNS randomly pick a few words to act as the ``loser" words, so that the winner word $w_O$ is differentiated \cite{rothe2015}.

As a result, by the gradient ascent updating, the input vector $v_{w_s}$ will gradually match with the output vectors $v_{w}'$ of words $w$ that appear in the context of $w_s$, while differ from output vectors $v_{\tilde{w}}'$ of all the words $\tilde{w}$ that are not in the context of $w_s$. In this way, the semantic information is therefore captured by the distribution of embedding vectors in the vector space \cite{turney2010} \cite{turney2013}.

Similarly, for an output vector $v_{w_s}^{\prime}$, the gradient of $E$ on each of its dimension is given as:
\begin{align*}
 & \frac{\partial E}{\partial v_{w_{s,i}}'} =
  \frac{1}{T}\sum_{t=1}^{T}\left(\!\!\left(\!\!\sum_{\mbox{\scriptsize$\begin{array}{c}-c\le j \le c,\\
j \ne 0,w_{t+j}=w_s
\end{array}$}}\!\!\!\!\!\!
  v_{w_{t,i}}\!\right)-2c\frac{\exp(v_{w_s}'^T v_{w_{t}})v_{w_{t,i}}}
  {\sum_{\tilde{w}=1}^{W}\exp(v_{\tilde{w}}'^T v_{w_t})}\!\!\right)\\
  & = \frac{1}{T}\sum_{t=1}^{T}\Big(n_{t,c,w_s}\cdot v_{w_{t,i}}-2c\cdot\hat{p}(w_s|w_t)\cdot v_{w_{t,i}}\Big)\\
  & = \frac{1}{T}\sum_{t=1}^{T}\Big(n_{t,c,w_s}-2c\cdot\hat{p}(w_s|w_t)\Big)v_{w_{t,i}}
\end{align*}
where $n_{t,c,w_s}$ means the number of times the word $w_s$ appears in the radius-c, center-removed window at the word $w_t$. And similar to the gradient of the input vector $v_{w_s}$, $\hat{p}(w_s|w_t)$ is the estimation of the conditional probability $p(w_s|w_t)$ given by the current vector set $\{v_w,v_w'\}_{w=1}^W$.

Then, the gradient formula of the entire output vector $v_{w_s}'$ is:
$$\frac{\partial E}{\partial v_{w_{s}}'} = \frac{1}{T}\sum_{t=1}^{T}\Big(n_{t,c,w_s}-2c\cdot\hat{p}(w_s|w_t)\Big)v_{w_{t}}.$$

And therefore, the gradient ascent updating rule for the output vector $v_{w_s}'$ will be:
\begin{align}
v_{w_{s}}'^{\text{ new}} = v_{w_{s}}'^{\text{ old}} + \frac{\eta}{T}\sum_{t=1}^{T}\Big(n_{t,c,w_s}-2c\cdot\hat{p}(w_s|w_t)\Big)v_{w_{t}}.
\end{align}

By looking at formula (5), it is easy to see that the softmax definition $\hat{p}(w_s|w_t) = \exp(v_{w_s}'^Tv_{w_t})/\big(\sum_{\tilde{w}=1}^{W}\exp(v_{\tilde{w}}'^T v_{w_t})\big)$ shall make $\hat{p}(w_s|w_t)$ small (less than $10^{-3}$) for most word pair $(w_s, w_t)$. And the window size $c$ is usually around $3$ to $5$, which means that multiplying $2c$ will not significantly enlarge $2c\cdot\hat{p}(w_s|w_t)$. As a result, the term $n_{t,c,w_s}-2c\cdot\hat{p}(w_s|w_t)$ will almost always be positive when the word $w_s$ appears in the context window of $w_t$, since in that case $n_{t,c,w_s}$ will be at least one. But if $w_s$ is not in the context window of $w_t$, $n_{t,c,w_s}$ will be zero and hence $n_{t,c,w_s}-2c\cdot\hat{p}(w_s|w_t)$ will be negative. 

That is, if the word $w_t=w$ at position $t$ in the training corpus has $w_s$ in its context window, it shall ``pull" the output vector $v_{w_s}'$ of the word $w_s$ towards its own input vector $v_{w_t}=v_w$. However, if $w_s$ is not in the context window of $w_t$, then the word $w_t=w$ has to ``push" $v_{w_s}'$ away from its own input vector $v_w$. Once again, this is a process of competitive learning: Each word $w_t=w$ in the training corpus shall compete against each other to make the output vector $v_{w_s}'$ move towards its own input vector $v_{w_t}=v_w$, otherwise the word $w_t=w$ at position $t$ will be competed out such that $v_{w_s}'$ will move away from $v_w$. Under this mechanism, those words with the word $w_s$ in their context window will win, and those without $w_s$ in their context window will lose. The competitive learning rule here is still ``winners take their winnings while losers even lose".

However, analyzing formula (3) again, we can see that in each training step the there is only one unique input vector $v_{w_I}$. So, the role played by $w_I$ need to be regarded as ``multiple": It is the ``winner" to the truly appeared word $w_O$, but it is the ``loser" to all the negative sampling words \cite{roman2018}. Hence, we have to admit that the competitive learning rule on updating the output vector of the skip-gram model is not fully reflected in the SGNS. In other words, the SGNS put a bias on the input vectors, while theoretically the status of input and output vectors in the skip-gram model should be equivalent \cite{socher2012}.

In summation, the gradient updating formulas of the input and output vectors in the skip-gram model are connected to the competitive learning rules, which are inherited by the SGNS. 

Based on the discussion in this section, we shall do analysis on the global optimal solution constraints of the skip-gram model in Section IV.\\

\section{Optimal Solutions for the Skip-Gram Model}
In this section, we shall first provide the global optimal solution constraints on the skip-gram model. Then, we shall use experimental results to support our analysis. Then, we will do analysis to show how the gradient ascent formulas of the skip-gram model make the word embedding vectors converge to the global optimal solution.

\subsection{The Global Optimal Solution Constraints}
While the gradient ascent formulas in the previous section result in the learning rule for one-step updating, it is desirable to know the properties of the global optimal solutions for quantitative analysis and reasoning: We concern about not only the rules to update input and output vectors, but also the final results of them. 

To do this, we will first reorder the terms by putting together all the context words for each word. That is, we will reform $E$ as:
\begin{align*}
  E & = \frac{1}{T}\sum_{t=1}^{T}\sum_{-c\le j \le c, j \ne 0}\log \hat{p}(w_{t+j}|w_t)\\
  & = \frac{1}{T}\sum_{w_s=1}^{W}\sum_{w=1}^{W}n_{w_s,w}\cdot \log \hat{p}(w|w_{s})\\
  & = \frac{1}{T}\sum_{w_s=1}^{W}\log\left(\prod_{w=1}^W\hat{p}(w|w_{s})^{n_{w_s,w}}\right)
\end{align*}

Here, $n_{w_s,w}$ is the number of times word $w$ appears in the radius-c, center-removed window of word $w_s$ throughout the training corpus (the counting is overlapping: if at time step $t$, $w_t=w$ appears in two overlapping windows of two nearby words, then $w_t=w$ will be counted twice). Then, the global optimal problem of the skip-gram model can be defined as: 

Given the word training corpus $\{w_1,\cdots,w_T\}$ and its corresponding dictionary $\mathcal{D}=\{w_1,\cdots,w_W\}$, we want to find an input and output vector set $\{v_w,v_w'\}_{w=1}^W$ with respect to each word $w$ in the dictionary $\mathcal{D}$, such that under the definition
$$\hat{p}(w|w_s)=\frac{\exp({v_{w}'}^Tv_{w_s})}{\sum_{\tilde{w}=1}^W\exp({v_{\tilde{w}}'}^Tv_{w_s})}$$
for any two words $w,w_s\in \mathcal{D}$, the average log probability 
\begin{align}
  E = \frac{1}{T}\sum_{w_s=1}^{W}\log\left(\prod_{w=1}^W\hat{p}(w|w_{s})^{n_{w_s,w}}\right)
\end{align}
is maximized.

Maximizing formula (6) directly is difficult. So, we are seeking for maximizing each term in it. That is, given a fixed training corpus $\{w_1,\cdots,w_T\}$ (and hence a fixed dictionary $\mathcal{D}=\{w_1,\cdots,w_W\}$ and fixed $n_{w_s,w}$ for $w=1,\cdots,W$), we want to maximize $\prod_{w=1}^W\hat{p}(w|w_{s})^{n_{w_s,w}}$ for each word $w_s\in \mathcal{D}$.

By the definition of $\hat{p}(w|w_s)$, we can see that no matter what the vector set is, $0\leq\hat{p}(w|w_s)\leq 1$ is always true for any two words $w$, $w_s$; and $\sum_{w=1}^W\hat{p}(w|w_s)=1$ is always true for any word $w_s$.

Therefore, the problem to maximizing $\prod_{w=1}^W\hat{p}(w|w_{s})^{n_{w_s,w}}$ for each specific word $w_s$ is equivalent to a constrained optimization problem that can be directly solved by the method of Lagrange Multiplier: Given a set of integers $\{n_1,n_2,\cdots ,n_W\}$ ($n_w=n_{w_s,w}$ for $w=1,\cdots,W$), we want to find $W$ non-negative real numbers $\{p_1,p_2,\cdots,p_W\}$ with the constraint $g(p_1,p_2,\cdots,p_W)=1-\sum_{w=1}^Wp_w=0$, so that the product  $$f(p_1,p_2,\cdots,p_W)=p_1^{n_1}p_2^{n_2}\cdots p_W^{n_W}$$
is maximized.

Now, define the Lagrange function to be: 
$$\mathcal{L}(p_1,p_2,\cdots,p_W,\lambda)=f(p_1,p_2,\cdots,p_W)+\lambda g(p_1,p_2,\cdots,p_W).$$ 

According to the method of Lagrange Multiplier, if $\{\hat{p}_1,\hat{p}_2,\cdots,\hat{p}_W\}$ is a solution to the initial optimization problem with constraint, then there exists a $\hat{\lambda}$ such that $$\nabla_{p_1,\cdots,p_W,\lambda}\mathcal{L}(p_1,p_2,\cdots,p_W,\lambda)=0$$
at $(\hat{p}_1,\hat{p}_2,\cdots,\hat{p}_W,\hat{\lambda})$.

Solving the above formula, we obtain a system of equations with $W+1$ equations and $W+1$ variables:
$$\begin{cases}
\left(n_1p_1^{n_1-1}\right)p_2^{n_2}\cdots p_W^{n_W}-\lambda=0\\
p_1^{n_1}\left(n_2p_2^{n_2-1}\right)\cdots p_W^{n_W}-\lambda=0\\
 \ \ \ \ \ \ \ \ \  \ \vdots\\
p_1^{n_1}p_2^{n_2}\cdots \left(n_Wp_W^{n_W-1}\right)-\lambda=0\\
1-\sum_{w=1}^Wp_w=0
\end{cases}$$

Taking the first two equations, we can get that 
$$\left(n_1p_1^{n_1-1}\right)p_2^{n_2}\cdots p_W^{n_W}=p_1^{n_1}\left(n_2p_2^{n_2-1}\right)\cdots p_W^{n_W}=\lambda.$$

Dividing by the common factors, we can obtain that $\left(n_1p_1^{n_1-1}\right)p_2^{n_2}=p_1^{n_1}\left(n_2p_2^{n_2-1}\right)$, which indicate that $n_1p_2=n_2p_1$. That is, $p_1:p_2=n_1:n_2$, which can be directly generalized as $p_w:p_u=n_w:n_u$ for any $w,u=1,\cdots,W$. Therefore, we can obtain that $p_1:p_2:\cdots :p_W=n_1:n_2:\cdots :n_W$. Finally, taking the last equation $1-\sum_{w=1}^Wp_w=0$, we can get that $$p_w=\frac{n_w}{\big(\sum_{\tilde{w}=1}^Wn_{\tilde{w}}\big)} \ \ \ \mathrm{for} \  w=1,\cdots,W.$$

But taking each $n_w$ back to $n_{w_s,w}$, we can easily see that 
$$\frac{n_w}{\big(\sum_{\tilde{w}=1}^Wn_{\tilde{w}}\big)}=\frac{n_{w_s,w}}{\big(\sum_{\tilde{w}=1}^Wn_{w_s,\tilde{w}}\big)}=\frac{n_{w_s,w}}{2c\cdot n_{w_s}},$$
where $n_{w_s}$ is the number of times the word $w_s$ appears in the training corpus. Now, notice that the term $n_{w_s,w}/\big(2c\cdot n_{w_s}\big)$ means: the number of times word $w$ appears in the context window of the word $w_s$ over the total amount of context the word $w_s$ has. That is, the probability of the word $w$ appears in the radius-c context window of the word $w_s$, which can be regarded as a ground-truth probability decided by the training corpus. We use $p_c(w|w_s)$ to represent this probability.

 As a result, we obtained the solution on maximizing $\prod_{w=1}^W\hat{p}(w|w_{s})^{n_{w_s,w}}$ for each word $w_s\in \mathcal{D}$. That is, we want the vector set satisfices:
\begin{align*}
\hat{p}(w|w_s)=\frac{\exp({v_{w}'}^Tv_{w_s})}{\sum_{\tilde{w}=1}^W\exp({v_{\tilde{w}}'}^Tv_{w_s})}=\frac{n_{w_s,w}}{2c\cdot n_{w_s}}=p_c(w|w_s).
\end{align*}

This means that: In order to maximize $\prod_{w=1}^W\hat{p}(w|w_{s})^{n_{w_s,w}}$ for each word $w_s$, the input and output vectors should make the estimated probability $\hat{p}(w|w_s)$ coincide with the ground true probability $p_c(w|w_s)$ for all the word $w$ in the dictionary.

As a result, here we provide our conclusion on the global optimal solution of the skip-gram model with word2vec:

Given the word training corpus $\{w_1,\cdots,w_T\}$ and its corresponding dictionary $\mathcal{D}=\{w_1,\cdots,w_W\}$, the average log probability $$E=\frac{1}{T}\sum_{t=1}^{T}\sum_{-c\le j \le c, j \ne 0}\log \hat{p}(w_{t+j}|w_t)$$ of the skip-gram model is maximized, when the input and output vector set $\{v_w,v_w'\}_{w=1}^W$ makes the estimated probability $\hat{p}(w_O|w_I)$ equal to the ground-truth probability $p_c(w_O|w_I)$ for any two words $w_I$, $w_O$ in the dictionary. That is, 
\begin{align}
\hat{p}(w_O|w_I)=\frac{\exp({v_{w_O}'}^Tv_{w_I})}{\sum_{w=1}^W\exp({v_{w}'}^Tv_{w_I})}=\frac{n_{w_I,w_O}}{2c\cdot n_{w_I}}=p_c(w_O|w_I)
\end{align}
for any $w_I, w_O\in \mathcal{D}$, where $n_{w_I,w_O}$ is the number of times the word $w_O$ appears in the radius-c, center-removed window of the word $w_I$ throughout the training corpus, and $n_{w_I}$ is the number of times the word $w_I$ appears in the training corpus.

However, note that formula (7) is only a constraint based on the inner products between input and output vectors. It does not specify the exact positions in the vector space, which is somehow typical in many optimization projects \cite{turian2010} \cite{Saadat2017}. Actually, there are infinite number of vector sets satisfying formula (7): If $V=\{v_w,v_w'\}_{w=1}^W$ satisfies formula (7), then any rotation of $V$ also does. The specific vector set $V$ obtained after training depends on the initial condition of vectors.

\subsection{Experimental Results}
In this subsection, we shall first provide the experimental results obtained on a toy training corpus, the words of the song $little \ star$, to support our analysis on the global optimal solution constraints. 

This song goes like: 
\textsl{``Every person had a star, every star had a friend, and for every person carrying
a star there was someone else who reflected it, and everyone
carried this reflection like a secret confidante in the heart." } Based on this toy corpus, we will strictly implement formula (1) to compute $\hat{p}(w_O|w_I)$ in the skip-gram model. We set $c=2$ and then go over the song for 500 times to maximize formula (2). 

After that, taking the word ``every" as an example, we look at both the ground-truth probability $p(w|``every")$ and the estimated one $\hat{p}(w|``every")$ for all the word $w$ appeared in the corpus: We can see that the word ``every" appears three times in the corpus. Hence, $n_{every}=3$ and $2c\cdot n_{every}=12.$ Then, we just count $n_{every,w}$ for each word $w$ in order to get $p(w|``every")=n_{every,w}/(2c\cdot n_{every})$. After that, we read out all the trained vectors $v_w$, $v_w'$ for each word $w$ to compute $\hat{p}(w|``every")$ based on formula (1). The result is in Table I:

\begin{table}[htbp]
\caption{The ground-truth probability and estimated probability for the skip-gram model trained on the toy corpus $LITTLE \ STAR$.}
\begin{center}
\begin{tabular}{|c|c|c||c|c|c|}
\hline
\!\!\!\text{Word $w$}\!\!\!&$\!\!\!\!\!p(w|every)\!\!\!\!\!$&$\!\!\!\!\!\hat{p}(w|every)\!\!\!\!\!$&\!\!\!\!\!\text{Word $w$}\!\!\!\!\!&$\!\!\!\!\!p(w|every)\!\!\!\!\!$&$\!\!\!\!\!\hat{p}(w|every)\!\!\!\!\!$\\
\hline
star &  0.1667 & 0.1718&friend &0.0000 & 0.0095\\
had & 0.1667 & 0.1713&reflection &0.0000 & 0.0092\\
person & 0.1667 & 0.1644&it &0.0000 & 0.0068\\
and & 0.0833 & 0.0917&secret &0.0000 & 0.0062\\
a & 0.0833 & 0.0893&was &0.0000 & 0.0061\\
for & 0.0833 & 0.0886&like &0.0000 & 0.0052\\
carrying & 0.0833 & 0.0865&everyone &0.0000 & 0.0046\\
every & 0.0000 & 0.0137&confidance &0.0000 & 0.0046\\
this & 0.0000 & 0.0109&heart &0.0000 & 0.0046\\
the &0.0000 & 0.0105&who &0.0000 & 0.0043\\
there &0.0000 & 0.0103&someone &0.0000 & 0.0043\\
else &0.0000 & 0.0100&carried &0.0000 & 0.0040\\
in &0.0000 & 0.0096&reflected &0.0000 & 0.0018\\
\hline
\end{tabular}

\label{tab1}
\end{center}
\end{table} 

And we provide the graph of $p(w|``every")$ and $\hat{p}(w|``every")$ with respect to words $w$ ordered as in above:
\begin{figure}[H]
\centering
\includegraphics[width=7cm]{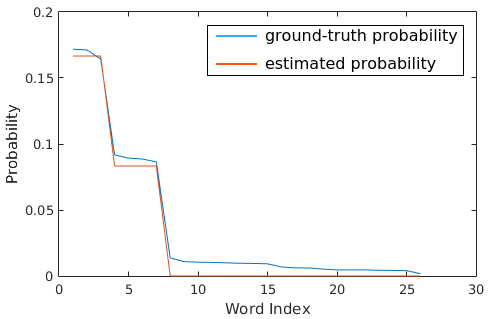}
\caption{The ground-truth and estimated probabilities based on the word ``every".}
\end{figure}

Based on the table and the graph, we can see that after training, the input and output vectors indeed converge to a stage satisfying the constraints of global optimal solution.

Then, we shall provide our experimental results obtained from a big training corpus. We use an optimized word2vec code provided online by the TensorFlow group \cite{tensorflow}. Our training corpus is dataset Text8, which consists of articles in English Wikipedia \cite{mahoney}. 

After the vector set $\{v_w,v_w'\}_{w=1}^W$ is trained, we shall choose a specific word $w_s$ to compute $p(w|w_s)$ and $\hat{p}(w|w_s)$ for the first 10,000 most frequent words $w$ in the dictionary. We use $S_{w_s}:\{p_{u|w_s}=p(w_u|w_s)\}_{u=1}^{10000}$ and $\hat{S}_{w_s}: \{\hat{p}_{u|w_s}=\hat{p}(w_u|w_s)\}_{u=1}^{10000}$ to represent the value set we obtained.
 
Then, regarding $S_{w_s}$ and $\hat{S}_{w_s}$ as two sample sets, we shall compute the correlation coefficient between the samples in them. That is:
\begin{align}
corr_{w_s}=\frac{\sum_{u=1}^{10000}(p_{u|w_s}-\Bar{p}_{w_s})(\hat{p}_{u|w_s}-\Bar{\hat{p}}_{w_s})}{\sqrt{\sum_{u=1}^{10000}(p_{u|w_s}-\bar{p}_{w_s})^2\cdot\sum_{u=1}^{10000}(\hat{p}_{u|w_s}-\bar{\hat{p}}_{w_s})^2}},
\end{align}
where $\bar{p}_{w_s}=\frac{\sum_{u=1}^{10000}p_{u|w_s}}{10000}$ and $\bar{\hat{p}}_{w_s}=\frac{\sum_{u=1}^{10000}\hat{p}_{u|w_s}}{10000}$ are the means of the samples in $S_{w_s}$ and $\hat{S}_{w_s}$ respectively.

The reason for us to calculate the correlation in such a way is that, the difference between $p(w_O|w_I)$ and $\hat{p}(w_O|w_I)$ for each pair of word $(w_I,w_O)$ appears to be chaotic in our experiments. We believe that the complexity of the big-data and the stochastic ambiguity caused by negative sampling shall generate noise upon the mathematical regularities, which makes the magnitude of $p(w_O|w_I)$, $\hat{p}(w_O|w_I)$ and $p(w_O|w_I)-\hat{p}(w_O|w_I)$ move out of their initial ratio. Hence, statistical methods are required to capture the relationship between $p(w_O|w_I)$ and $\hat{p}(w_O|w_I)$. We fix the input word $w_I = w_s$ since we can then make all the estimated probabilities $\hat{p}(w_u|w_s)$ share the same denominator $\sum_{w=1}^W\exp(v_{w}'^Tv_{w_s})$.

We choose 18 specific words as $w_s$, including 6 nouns, 6 verbs and 6 adjectives. The results on their correlation coefficients are shown in Table II: 
\begin{table}[htbp]
\caption{The correlation coefficients between the ground-truth probability and the estimated probability of each word $w_s$}
\begin{center}
\begin{tabular}{|c|c||c|c||c|c|}
\hline
\!\text{Word $w_s$}\!&$\!corr_{w_s}\!$&\!\text{Word $w_s$}\!&$\!corr_{w_s}\!$&\!\text{Word $w_s$}\!&$\!corr_{w_s}\!$\\
\hline
water&0.3558&run&0.3433&smart&0.3327\\
man&0.3230&play&0.3125&pretty&0.4039\\
king&0.3169&eat&0.3879&beautiful&0.3074\\
car&0.3300&drink&0.3507&dark&0.3209\\
bird&0.2700&fly&0.2886&high&0.3859\\
war&0.3990&draw&0.2730&low&0.3707\\
\hline
\end{tabular}

\label{tab1}
\end{center}
\end{table}

Since there are 10000 $w_u$ participating in the computation of $corr_{w_s}$ for each $w_s$, a correlation coefficient around 0.3 to 0.4 is significant. That is, for a fixed word $w_s$, there exists a linear relationship between the ground-truth probability $p(w_u|w_s)$ and the estimated probability $\hat{p}(w_u|w_s)$. And hence, the linear correlation: $p(w_O|w_I)=a\cdot\hat{p}(w_O|w_I)+b$ can be seen from the ambiguity of $w_s$, which is a strong support to our stated formula (7) in subsection A.

Based on the results from both the toy corpus and the big dataset, we can see that our global optimal solution constraints on the vectors in word2vec is correct.\footnote{Our work are open-source. Researchers can find our codes and datasets on our website: https://github.com/canlinzhang/IJCNN-2019-paper.}

\subsection{Connections to the Gradient Ascent Formulas}
Based on our results so far, researchers may ask whether the vectors would truly converge to the optimal solution under the gradient ascent formulas. The answer is yes and we shall show it in this subsection.

According to the gradient formula of the input vector $v_{w_s}$ in Section III, we can furthermore obtain that:

\begin{align*}
& \frac{\partial E}{\partial v_{w_{s}}} = \frac{1}{T}\!\!\!\!\!\sum_{\mbox{\scriptsize$\begin{array}{c}t=1,w_t=w_s 
\end{array}$}}^T\!\!\!\!\!\!\!\sum_{\mbox{\scriptsize$\begin{array}{c}-c\le j \le c, j \ne 0
\end{array}$}}\!\!\!\!\!\!
        \left(\!\!v_{w_{t+j}}'\!-\!\!\sum_{w=1}^{W}\hat{p}(w|w_s) v_{w}'\!\!\right)\\
        & = \frac{1}{T}\!\!\!\!\!\sum_{\mbox{\scriptsize$\begin{array}{c}t=1,w_t=w_s 
\end{array}$}}^T\!\!\!\!\!\!\left(\!\!\!\left(\!\!\!\!\!\sum_{\mbox{\scriptsize$\begin{array}{c}-c\le j \le c, j \ne 0
\end{array}$}}\!\!\!\!v_{w_{t+j}}'\!\!\right)\!-\! 2c\left(\sum_{w=1}^{W}\hat{p}(w|w_s) v_{w}'\!\right)\!\!\!\right)\\
& =\!\! \frac{1}{T}\!\!\left(\!\!\!\left(\!\!\!\!\sum_{\mbox{\scriptsize$\begin{array}{c}t=1,w_t=w_s 
\end{array}$}}^T\!\!\!\!\!\!\!\sum_{\mbox{\scriptsize$\begin{array}{c}-c\le j \le c,j \ne 0
\end{array}$}}\!\!\!\!\!\!\!v_{w_{t+j}}'\!\!\right)\!\!-\!\!2c n_{w_s}\!\left(\!\sum_{w=1}^{W}\!\!\hat{p}(w|w_s) v_{w}'\!\right)\!\!\!\right)
\end{align*}
Setting $n_{w_s,w}$ and $n_{w_s}$ to have the same meaning as in subsection A, we can get that:
\begin{align*}
\frac{\partial E}{\partial v_{w_{s}}} & = \frac{1}{T}\left(\!\!\!\left(\sum_{w=1}^Wn_{w_s,w}\cdot v_{w}'\right)-\left(\sum_{w=1}^{W}2c \cdot n_{w_s}\cdot \hat{p}(w|w_s) v_{w}'\right)\!\!\!\right)\\
& =\frac{1}{T} \sum_{w=1}^W\left(n_{w_s,w}-2c\cdot n_{w_s}\cdot\hat{p}(w|w_s)\right)v_{w}'\\
& =2c\cdot\frac{n_{w_s}}{T}\cdot\sum_{w=1}^W\left(\frac{n_{w_s,w}}{2c\cdot n_{w_s}}-\hat{p}(w|w_s)\right)\! v_{w}'
\end{align*}

Note that $\frac{n_{w_s}}{T}$ is actually the ground-truth occurring probability of the word $w_s$ based on the training corpus, which we denote as $p(w_s)$. Also, taking $\frac{n_{w_s,w}}{2c\cdot n_{w_s}}=p(w|w_s)$ as in subsection A, we have that:
$$\frac{\partial E}{\partial v_{w_{s}}}=2c\cdot p(w_s)\cdot\sum_{w=1}^W\Big(p(w|w_s)-\hat{p}(w|w_s)\Big) v_{w}'$$

Therefore, the updating rule of the input vector $v_{w_s}$ under this gradient ascent formula will be:
\begin{align}
v_{w_{s}}^{\text{new}}\! = v_{w_{s}}^{\text{old}}\! +\!\eta\!\cdot\! 2c p(w_s)\!\cdot\!\! \left(\!\sum_{w=1}^W\!\!\Big(p(w|w_s)-\hat{p}(w|w_s)\!\Big) v_{w}'\!\!\right)
\end{align}

Intuitively speaking, this formula shows that in the skip-gram model, the wider the context window is, the faster all the input vectors will change. And the more frequent a word is, the fast its input vector may change. However, the essential part of formula (9) is the summation $\sum_{w=1}^W\Big(p(w|w_s)-\hat{p}(w|w_s)\Big) v_{w}'$, in which lies the connection between the gradient ascent rule of the input vector and the global optimal solution constraints.  

For any two fixed words $w$ and $w_s$, we have that
\begin{align*}
 v_w'^T & v_{w_{s}}^{\text{new}}\!=\!v_w'^T \!\Big[\! v_{w_{s}}^{\text{old}}\! +\!\eta\cdot 2c p(w_s)\cdot\! \Big(\!\!\sum_{\tilde{w}=1}^W\!\!\Big(\!p(\tilde{w}|w_s)\!-\!\hat{p}(\tilde{w}|w_s)\!\Big)\! v_{\tilde{w}}'\!\Big)\!\Big]\\
= & v_w'^T\cdot\Big[v_{w_{s}}^{\text{old}} + 2c\eta p(w_s)\cdot \Big(p(w|w_s)-\hat{p}(w|w_s)\Big)v_w'\\
& \ \ \ \ \ \ \ \ \ \ \  +  2c\eta p(w_s)\cdot\!\!\!\!\!\!\!\sum_{\mbox{\scriptsize$\begin{array}{c}\tilde{w}=1,\tilde{w}\neq w
\end{array}$}}^W\!\!\!\!\!\!\!\Big(p(\tilde{w}|w_s)-\hat{p}(\tilde{w}|w_s)\Big)v_{\tilde{w}}'\Big]\\
= & v_w'^Tv_{w_{s}}^{\text{old}}+ 2c\eta p(w_s)\cdot\Big(p(w|w_s)-\hat{p}(w|w_s)\Big)|\!|v_w'|\!|_2^2\\
& \ \ \ \ \ \ \ \  +  2c\eta p(w_s)\cdot\!\!\!\!\!\!\sum_{\mbox{\scriptsize$\begin{array}{c}\tilde{w}=1,\tilde{w}\neq w
\end{array}$}}^W\!\!\!\!\!\!\!\Big(p(\tilde{w}|w_s)-\hat{p}(\tilde{w}|w_s)\Big)v_w'^Tv_{\tilde{w}}'.
\end{align*}

It is easy to see that $|v_w'^Tv_{\tilde{w}}'|=\Big|\sum_{i=1}^Kv_{w_i}'v_{\tilde{w}_i}'\Big|\ll |\!|v_w'|\!|_2^2$ is true for almost all words $\tilde{w}$. This is because the two output vectors $v_w'$ and $v_{\tilde{w}}'$ can hardly be in the same direction of the vector space. As a result, terms in the summation  $\sum_{i=1}^Kv_{w_i}'v_{\tilde{w}_i}'$ may largely cancel each other. Besides, we claim that the sign of $(p(\tilde{w}|w_s)-\hat{p}(\tilde{w}|w_s))$, the magnitude of $(p(\tilde{w}|w_s)-\hat{p}(\tilde{w}|w_s))$ and the sign of $v_w'^Tv_{\tilde{w}}'$ should be independent with respect to each other due to the complexity of words' contexts and the random initialization of vectors. This means that terms in the summation $\sum_{\tilde{w}=1,\tilde{w}\neq w}^W\!\!\Big(p(\tilde{w}|w_s)-\hat{p}(\tilde{w}|w_s)\Big)v_w'^Tv_{\tilde{w}}'$ will not accumulate towards positive (or negative). Instead, these terms will largely cancel each other through the summation.

As a result, when the vector set $\{v_w,v_w'\}_{w=1}^W$ significantly under-estimates the conditional probability $p(w|w_s)$ (say, $p(w|w_s)-\hat{p}(w|w_s)>0.5p(w|w_s)$), we can infer that the term $\Big(p(w|w_s)-\hat{p}(w|w_s)\Big)|\!|v_w'|\!|_2^2$ is significantly larger than $\sum_{\mbox{\scriptsize$\begin{array}{c}\tilde{w}=1,\\
 \tilde{w}\neq w
\end{array}$}}^W\!\!\!\!\!\!\!\Big(p(\tilde{w}|w_s)-\hat{p}(\tilde{w}|w_s)\Big)v_w'^Tv_{\tilde{w}}'$. This means that $v_w'^T v_{w_{s}}^{\text{new}}> v_w'^T v_{w_{s}}^{\text{old}}$, and hence $\exp( v_w'^T v_{w_{s}}^{\text{new}})>\exp( v_w'^T v_{w_{s}}^{\text{old}})$. That is, the numerator of $\hat{p}(w|w_s)$ will increase after the gradient ascent updating of $v_{w_s}$.

On the other hand, looking at the denominator $\sum_{\tilde{w}=1}^W\exp(v_{\tilde{w}}'^T v_{w_{s}})$ of the term $\hat{p}(w|w_s)$, one can see that due to the complexity of words' contexts and the random initialization of vectors, we can get two statements: First, the number of words $\tilde{w}$ with $\exp(v_{\tilde{w}}'^T v_{w_{s}}^{\text{new}})>\exp(v_{\tilde{w}}'^T v_{w_{s}}^{\text{old}})$ shall mostly equals to that with $\exp(v_{\tilde{w}}'^T v_{w_{s}}^{\text{new}})<\exp(v_{\tilde{w}}'^T v_{w_{s}}^{\text{old}})$. Second, the magnitude of $\big(\exp(v_{\tilde{w}}'^T v_{w_{s}}^{\text{new}})-\exp(v_{\tilde{w}}'^T v_{w_{s}}^{\text{old}})\big)$ should be independent to the its sign. That is, the difference in each term $\exp(v_{\tilde{w}}'^T v_{w_{s}})$ caused by updating $v_{w_s}$ should largely be cancelled by the summation $\sum_{\tilde{w}=1}^W\exp(v_{\tilde{w}}'^T v_{w_{s}})$. Hence, we can see that the denominator of $\hat{p}(w|w_s)$ will usually not have an obvious change by one step updating on $v_{w_s}$.

Based on the above analysis, we may conclude that when $\hat{p}(w|w_s)$ is significantly smaller than $p(w|w_s)$, the new $\hat{p}(w|w_s)$ shall increase after one step updating on $v_{w_s}$, assuming all other vectors are invariant. Similarly, when $\hat{p}(w|w_s)$ is significantly larger than $p(w|w_s)$, the new $\hat{p}(w|w_s)$ shall decrease after updating the input vector $v_{w_s}$. That is, the gradient ascent rule of the input vector $v_{w_s}$ provides an auto-adjusting mechanism to make sure that the estimation probability $\hat{p}(w|w_s)$ shall gradually converge to the ground-truth probability $p(w|w_s)$ in the training, which also makes the input vector $v_{w_s}$ converge to a stage satisfying the global optimal solution constraints \cite{collobert2008}.

For an output vector $v_{w_s}^{\prime}$, the gradient is given as:
\begin{align*}
 \frac{\partial E}{\partial v_{w_{s,i}'}} =&
  \frac{1}{T}\sum_{t=1}^{T}\left(\!\!\left(\!\!\sum_{\mbox{\scriptsize$\begin{array}{c}-c\le j \le c,\\
j \ne 0,w_{t+j}=w_s
\end{array}$}}\!\!\!\!\!
  v_{w_{t,i}}\!\right)-2c\frac{\exp(v_{w_s}'^T v_{w_{t}})v_{w_{t,i}}}
  {\sum_{\hat{w}=1}^{W}\exp(v_{\hat{w}}'^T v_{w_t})}\!\!\right) \\
  =\frac{1}{T}&\left(\!\!\left(\!\!\sum_{t=1}^{T}\!\!\sum_{\mbox{\scriptsize$\begin{array}{c}-c\le j \le c,\\
j \ne 0,w_{t+j}=w_s
\end{array}$}}\!\!\!\!\!
  v_{w_{t,i}}\!\right)-2c\sum_{t=1}^T\frac{\exp(v_{w_s}'^T v_{w_{t}})v_{w_{t,i}}}
  {\sum_{\hat{w}=1}^{W}\exp(v_{\hat{w}}'^T v_{w_t})}\!\!\right). 
  \end{align*}

Looking at the summations on the first $v_{w_{t,i}}$, it means that if the word $w_s$ appears in the radius-$c$ window of $w_t$ in the training corpus, the $i$'th dimension of the vector $v_{w_{t}}$ of $w_t$ will be picked out for the summation. This explanation may be difficult to accept intuitively. But if we pay attention to each $w_{\tilde{t}}=w_s$ in the training corpus, we can figure out that all the words $w_t$ within the radius-$c$ window of $w_{\tilde{t}}$ will be picked out for the summation. As a result, this is exactly the same as the summation on $v_{t+j,i}'$ with respect to the two conditions $t=1, w_t=w_s$ and $-c\leq j\leq c, j\neq 0$ (Only here the summation is on the input vectors). That is:
$$\sum_{t=1}^{T}\!\!\sum_{\mbox{\scriptsize$\begin{array}{c}-c\le j \le c,\\
j \ne 0,w_{t+j}=w_s
\end{array}$}}\!\!\!
  v_{w_{t,i}}=\!\!\sum_{\mbox{\scriptsize$\begin{array}{c}t=1,\\
w_t=w_s 
\end{array}$}}^T\!\!\!\!\!\!\!\sum_{\mbox{\scriptsize$\begin{array}{c}-c\le j \le c,\\
 j \ne 0
\end{array}$}}\!\!\!v_{w_{t+j,i}}=\sum_{w=1}^Wn_{w_s,w}\cdot v_{w_i},$$
where $n_{w_s,w}$ is defined the same to be the number of times word $w$ appears in the window of $w_s$ throughout the corpus.

Now look at the second summation: Although it is a summation from $t=1$ to $T$, we shall figure out that each term in the summation actually does not depend on time step $t$. It only depends on the vector $v_w$ of the word $w$ at time step $t$. That is:
\begin{align*}
&\sum_{t=1}^T\frac{\exp(v_{w_s}'^T v_{w_{t}})v_{w_{t,i}}}
  {\sum_{\hat{w}=1}^{W}\exp(v_{\hat{w}}'^T v_{w_t})} =\!\!\!\!\! \sum_{\mbox{\scriptsize$\begin{array}{c}t=1,w_t=w
\end{array}$}}^T \!\!\!\! \frac{\exp(v_{w_s}'^T v_w)v_{w_i}}
  {\sum_{\hat{w}=1}^{W}\exp(v_{\hat{w}}'^T v_w)}\\
 & = \!\!\! \sum_{\mbox{\scriptsize$\begin{array}{c}t=1, w_t=w
\end{array}$}}^T \!\!\!\!\hat{p}(w_s|w)v_{w_i} =  \sum_{w=1}^W n_w\cdot\hat{p}(w_s|w)v_{w_i},
\end{align*}
where $n_w$ is the number of times the word $w$ appears in the training corpus.

Therefore, taking the above formulas into the gradient of the output vector $v_{w_s}'$ and applying similar concepts and tricks as computing the gradient of input vector $v_{w_s}$, we have that:
\begin{align*}
& \frac{\partial E}{\partial v_{w_{s,i}'}} = \frac{1}{T}\!\left(\!\!\left(\sum_{w=1}^Wn_{w_s,w}\cdot v_{w_i}\right)-2c\left(\sum_{w=1}^W n_w\cdot\hat{p}(w_s|w)v_{w_i}\right)\!\!\right)\\
& =\!\!  \frac{1}{T}\!\!\sum_{w=1}^W\!\!\Big(\!n_{w_s,w}\!-\!2c\!\cdot\! n_w\!\cdot \!\!\hat{p}(\!w_s|w\!)\!\!\Big)\!v_{w_i}\!\! =\!\!\!\!  \sum_{w=1}^W\!\!2c\!\cdot\!\frac{n_w}{T}\!\Big(\!\frac{n_{w_s,w}}{2c\cdot n_w}\!-\! \hat{p}(\!w_s|w\!)\!\!\Big)\!v_{w_i}\\
& =  \sum_{w=1}^W2c\cdot p(w)\Big(p(w_s|w)- \hat{p}(w_s|w)\Big)v_{w_i}
\end{align*}

Although it may appear to be straightforward if we multiply the term $p(w)$ into the conditional probabilities $p(w_s|w)$ and $\hat{p}(w_s|w)$, this multiplication will lead to $p(w)\cdot\hat{p}(w_s|w)$, which is a multiplication between the ground-truth probability and the estimated conditional probability, making no sense intuitively. As a result, we just keep the ground-truth probability $p(w)$ and obtain another gradient formula for the entire output vector $v_{w_s}'$:
\begin{align*}
\frac{\partial E}{\partial v_{w_s'}}= \sum_{w=1}^W2c\cdot p(w)\Big(p(w_s|w)- \hat{p}(w_s|w)\Big)v_{w}.
\end{align*}

Therefore, the gradient ascent updating rule of the output vector $v_{w_s}'$ will be:
\begin{align}
v_{w_{s}}'^{\text{new}} = v_{w_{s}}'^{\text{old}} + \eta\sum_{w=1}^W2c\cdot p(w)\Big(p(w_s|w)- \hat{p}(w_s|w)\Big)v_{w}.
\end{align}

Similarly to the analysis on the gradient ascent formula of the input vector $v_{w_s}$ in this subsection, we have that:
\begin{align*}
&{v_{w_{s}}'^{\text{new}}}^T v_{w}\!=\!\Big[v_{w_{s}}'^{\text{old}} + \eta\!\sum_{\tilde{w}=1}^W\!2c\cdot p(\tilde{w})\Big(p(w_s|\tilde{w})- \hat{p}(w_s|\tilde{w})\Big)v_{\tilde{w}}\Big]^T\!v_w\\
&={v_{w_{s}}'^{\text{old}}}^Tv_w + 2c\eta p(w)\Big(p(w_s|w)- \hat{p}(w_s|w)\Big)|\!|v_w|\!|_2^2\\
& \ \ \ \ \  \ \ \ \ \ \ \ \ +\!\sum_{\tilde{w}=1,\tilde{w}\neq w}^W\!2c\eta p(\tilde{w})\Big(p(w_s|\tilde{w})- \hat{p}(w_s|\tilde{w})\Big)v_{\tilde{w}}^Tv_w
\end{align*}

Once again, we can see that: $|v_{\tilde{w}}^Tv_w|\ll |\!|v_w|\!|_2^2$ due to the complexity of the words' context as well as the random initialization of the embedding vectors; And terms in the summation $\sum_{\tilde{w}=1}^W\!2c\eta p(\tilde{w})\Big(p(w_s|\tilde{w})- \hat{p}(w_s|\tilde{w})\Big)v_{\tilde{w}}^Tv_w$ will largely cancel each other due to the independencies among the sign of $\big(p(w_s|\tilde{w})- \hat{p}(w_s|\tilde{w})\big)$, the magnitude of $ p(\tilde{w})\Big(p(w_s|\tilde{w})- \hat{p}(w_s|\tilde{w})\Big)$ and the sign of $v_{\tilde{w}}^Tv_w$. However, note that the term $p(\tilde{w})$ is the ground-truth probability of each word $\tilde{w}$, which varies significantly from frequent to rare words. Hence, we can see that $2c\eta p(w)\Big(p(w_s|w)- \hat{p}(w_s|w)\Big)|\!|v_w|\!|_2^2>\sum_{\tilde{w}=1}^W\!2c\eta p(\tilde{w})\Big(p(w_s|\tilde{w})- \hat{p}(w_s|\tilde{w})\Big)v_{\tilde{w}}^Tv_w$ is more likely to be true when $p(w)$ is big and $\hat{p}(w_s|w)$ is significantly smaller than $p(w_s|w)$. For rare words $w$ with small $p(w)$, they have a chance to satisfy this claim after enough many steps of updating on the output vector $v_{w_s}'$. And our claim is equivalent to $\exp({v_{w_{s}}'^{\text{new}}}^T v_{w})>\exp({v_{w_{s}}'^{\text{old}}}^T v_{w})$.

Also, we can easily see that the denominator $\sum_{\tilde{w}=1}^Wv_{\tilde{w}}'v_w$ of $\hat{p}(w_s|w)$ will have no change after the updating on $v_{w_s}'$ since it is irrelative to $v_{w_s}'$ at all. As a result, we conclude that for any word $w$ and the fixed word $w_s$, after enough many steps of updating on the output vector $v_{w_s}'$, $\hat{p}(w_s|w)$ will increase if $\hat{p}(w_s|w)$ is significantly smaller than $p(w_s|w)$ at beginning. Similarly, $\hat{p}(w_s|w)$ will decrease after enough many steps of updating on the output vector $v_{w_s}'$, if $\hat{p}(w_s|w)$ is significantly larger than $p(w_s|w)$ at beginning. Again, this is an auto-adjusting mechanism making $v_{w_s}'$ converge to a stage satisfying the global optimal solution constraints.

After all, combining the adjusting mechanisms of the gradient ascent rules on input and output vectors, we can see that the vectors will dynamically works together to make the estimation probability $\hat{p}(w_O|w_I)$ converge to the ground-truth probability $p(w_O|w_I)$ for any word $w_I$, $w_O$ in the dictionary, which also means to converge to the global optimal solution of the skip-gram model.

\section{Conclusion and Future Work} 
In this work, we have provided a comprehensive mathematical analysis on the skip-gram model. We have derived the gradient ascent formulas for the input and output vectors, then connected these formulas to the competitive learning and the SGNS model. After that, we have provided the global optimal solution constraints on the vectors of the skip-gram model, with the support from experimental results. Finally, we have done analysis showing that under the gradient ascent formulas, the word embedding vectors of the skip-gram model will indeed converge to a stage satisfying the global optimal solution constraints.

In the future, we want to provide even deeper analysis showing quantitatively why the analogue relationships such as $v_{man}-v_{woman}\approx v_{king}-v_{queen}$ are satisfied by the word2vec vectors. Finally, based on the understanding of the state-of-the-art language model for deep learning, we hope to come up with a composite language model applying phrase embeddings, which can capture the meanings of phrases and semantic units \cite{rothe2015}.\\





\begin{thebibliography}{1}

\bibitem{bengio2003} Y. Bengio, R. Ducharme, P. Vincent, and C. Janvin, ``A neural probabilistic language
model,"
        {\it The Journal of Machine Learning Research}, no. 3, pp. 1137–1155, 2003.

\bibitem{collobert2008} R. Collobert and J. Weston, ``A unified architecture for natural language processing: deep neural networks with multitask learning,"
		{\it International Conference on Machine Learning}, pp. 160–167, 2008.
		
\bibitem{huang2012} E. Huang, R. Socher, C. Manning and A. Ng, ``Improving Word Representations via Global Context and Multiple Word Prototypes,"
		{\it Association for Computational Linguistics}, 2012.

\bibitem{initialize} T. Kocmi and O. Bojar, ``An Exploration of Word Embedding Initialization in Deep-Learning Tasks,"
		{\it arXiv:1711.09160v1 [cs.CL] 24 Nov 2017}, 2017.

\bibitem{mahoney} M. Mahoney, ``An introduction to the training corpus Text8,"
		{\it http://mattmahoney.net/dc/textdata.html}.
		
\bibitem{mikolov1} T. Mikolov, I. Sutskever, K. Chen, G. Corrado, and J. Dean ``Distributed representations of words and phrases and their compositionality,"
        {\it Neural Information Processing Systems}, 2013.
       
\bibitem{mikolov2} T. Mikolov, K. Chen, G. Corrado, and J. Dean ``Efficient estimation of word representations in vector space,"
        {\it arXiv preprint arXiv:1301.3781
, 2013a}. 

\bibitem{mikolov3} T. Mikolov, S. Kombrink, L. Burget, J. Cernocky, and S. Khudanpur, ``Extensions of recurrent neural network language model,"
		{\it Acoustics, Speech and Signal Processing (ICASSP), IEEE International Conference}, pp. 5528–5531, 2011.
		
\bibitem{mikolov4} T. Mikolov, A. Deoras, D. Povey, L. Burget and J. Cernocky, ``Strategies for Training Large Scale Neural Network Language Models,"
		{\it Automatic Speech Recognition and Understanding}, 2011. 
		
\bibitem{mikolov5} T. Mikolov, W. Yih and G. Zweig, ``Linguistic Regularities in Continuous Space Word Representations,"
		{\it North American Chapter of the Association for Computational Linguistics: Human Language Technologies (NAACL HLT)}, 2013. 
		
\bibitem{mnih2009} A. Mnih and G. E Hinton, ``A scalable hierarchical distributed language model,"
		{\it Advances in
neural information processing systems}, vol. 21, pp. 1081–1088, 2009. 

\bibitem{mnih2005} F. Morin and Y. Bengio, ``Hierarchical probabilistic neural network language model,"
		{\it International workshop on artificial intelligence and statistics, pages 246–252, 2005}, pp. 246-252, 2005.   
	

\bibitem{oja89} E. Oja, ``Neural networks, principle components, and subspaces,"
		{\it International Journal of Neural Systems}, vol. 1, pp. 61-68, 1989.  

\bibitem{roman2018} R. Roman, R. Precup and R. David, ``Second Order Intelligent Proportional-Integral Fuzzy Control of Twin Rotor Aerodynamic Systems,"
		{\it Procedia Computer Science}, vol. 139, pp. 372-380, 2018.

\bibitem{rothe2015} S. Rothe and H. Schütze, ``Autoextend: Extending word embeddings to embeddings for synsets and lexemes,"
		{\it Association for Computational Linguistics}, pp. 1793-1803, 2015.

\bibitem{Saadat2017} J. Saadat, P. Moallem and H. Koofigar, ``Training Echo State Neural Network Using Harmony Search Algorithm,"
        {\it  International Journal of Artificial Intelligence}, vol. 15, no. 1, pp. 163-179, 2017

\bibitem{shinozaki2018} T. Shinozaki, ``Competitive Learning Enriches Learning Representation and Accelerates the Fine-tuning of CNNs,"
		{\it arXiv:1804.09859v1 [cs.LG] 26 Apr 2018}, 2018.

\bibitem{socher2012} R. Socher, B. Huval, C. Manning, and A. Ng, ``Semantic Compositionality Through Recursive Matrix-Vector Spaces,"
		{\it Empirical Methods
in Natural Language Processing (EMNLP)}, 2012. 

\bibitem{tensorflow} TensorFlow group, ``Vector Representations of Words ,"
		{\it https://www.tensorflow.org/tutorials/representation/word2vec}. 

\bibitem{turian2010} J. Turian, L. Ratinov, and Y. Bengio, ``Word representations: a simple and general method for semi-supervised learning,"
		{\it Association for Computational Linguistics}, pp. 384–394, 2010.

\bibitem{turney2010} P. Turney and P. Pantel, ``From frequency to meaning: Vector space models of semantics,"
		{\it Journal of Artificial Intelligence Research}, vol. 37, pp. 141-188, 2010.  
		
\bibitem{turney2013} P. Turney, ``Distributional semantics beyond words: Supervised learning of analogy and paraphrase,"
		{\it Transactions of the Association for Computational Linguistics (TACL)}, pp. 353–366, 2013. 

	
\end{thebibliography}
%

\def\V{\rm vol.~}
\def\N{no.~}
\def\pp{pp.~}
\def\Pot{\it Proc. }
\def\IJCNN{\it International Joint Conference on Neural Networks\rm }
\def\ACC{\it American Control Conference\rm }
\def\SMC{\it IEEE Trans. Systems\rm , \it Man\rm , and \it Cybernetics\rm }

\def\handb{ \it Handbook of Intelligent Control: Neural\rm , \it
    Fuzzy\rm , \it and Adaptive Approaches \rm }

\end{document}